\documentclass[lettersize,journal]{IEEEtran}
\usepackage{lineno,hyperref}
\usepackage{amsmath,amsfonts}
\usepackage{array}
\usepackage{subfig}
\usepackage{textcomp}
\usepackage{stfloats}
\usepackage{url}
\usepackage{verbatim}
\usepackage{graphicx}
\usepackage{cite}
\usepackage{multirow}
\usepackage{bbding}
\usepackage{booktabs}
\usepackage{color}
\usepackage{xcolor}
\usepackage{colortbl}
\usepackage{amsfonts}
\usepackage{amsmath}
\usepackage[ruled,linesnumbered]{algorithm2e}
\usepackage{newfloat}
\usepackage{listings}

\def\eg{\emph{e.g.}} 
\def\ie{\emph{i.e.}} 
\def\etal{\emph{et~al.}} 
\newlength\savewidth\newcommand\shline{\noalign{\global\savewidth\arrayrulewidth
  \global\arrayrulewidth 1pt}\hline\noalign{\global\arrayrulewidth\savewidth}}

\begin{document}

\title{PiPa++: Towards Unification of Domain Adaptive Semantic Segmentation via Self-supervised Learning}

\author{Mu Chen, Zhedong Zheng, and Yi Yang,~\IEEEmembership{Senior Member, IEEE}
\thanks{
M. Chen is with ReLER Lab, AAII,  University of Technology Sydney, Australia. Z. Zheng is with FST and ICI, University of Macua, China. Y. Yang is with ReLER Lab, CCAI,  Zhejiang University, China.}

\thanks{
Corresponding author: Z. Zheng 
(zhedongzheng@um.edu.mo).}
}

\markboth{Journal of \LaTeX\ Class Files,~Vol.~18, No.~9, July~2024}%
{Shell \MakeLowercase{\textit{et al.}}: A Sample Article Using IEEEtran.cls for IEEE Journals}


\maketitle

\begin{abstract}
Unsupervised domain adaptive segmentation aims to improve the segmentation accuracy of models on target domains without relying on labeled data from those domains. This approach is crucial when labeled target domain data is scarce or unavailable. It seeks to align the feature representations of the source domain (where labeled data is available) and the target domain (where only unlabeled data is present), thus enabling the model to generalize well to the target domain. Current image- and video-level domain adaptation have been addressed using different and specialized frameworks, training strategies and optimizations despite their underlying connections. In this paper, we propose a unified framework PiPa++, which leverages the core idea of ``comparing'' to (1) explicitly encourage learning of discriminative pixel-wise features with intraclass compactness and inter-class separability, (2) promote the robust feature learning of the identical patch
against different contexts or fluctuations, and (3) enable the learning of temporal continuity under dynamic environments. With the designed task-smart contrastive sampling strategy, PiPa++ enables the mining of more informative training samples according to the task demand. Extensive experiments demonstrate the effectiveness of our method on both image-level and video-level domain adaption benchmarks. Moreover, the proposed method is compatible with other UDA approaches to further improve the performance without introducing extra parameters.

\end{abstract}


\section{Introduction}
\label{sec: Introduction}

\begin{figure}[t!]
    \centering
    \includegraphics[width=0.95\linewidth]{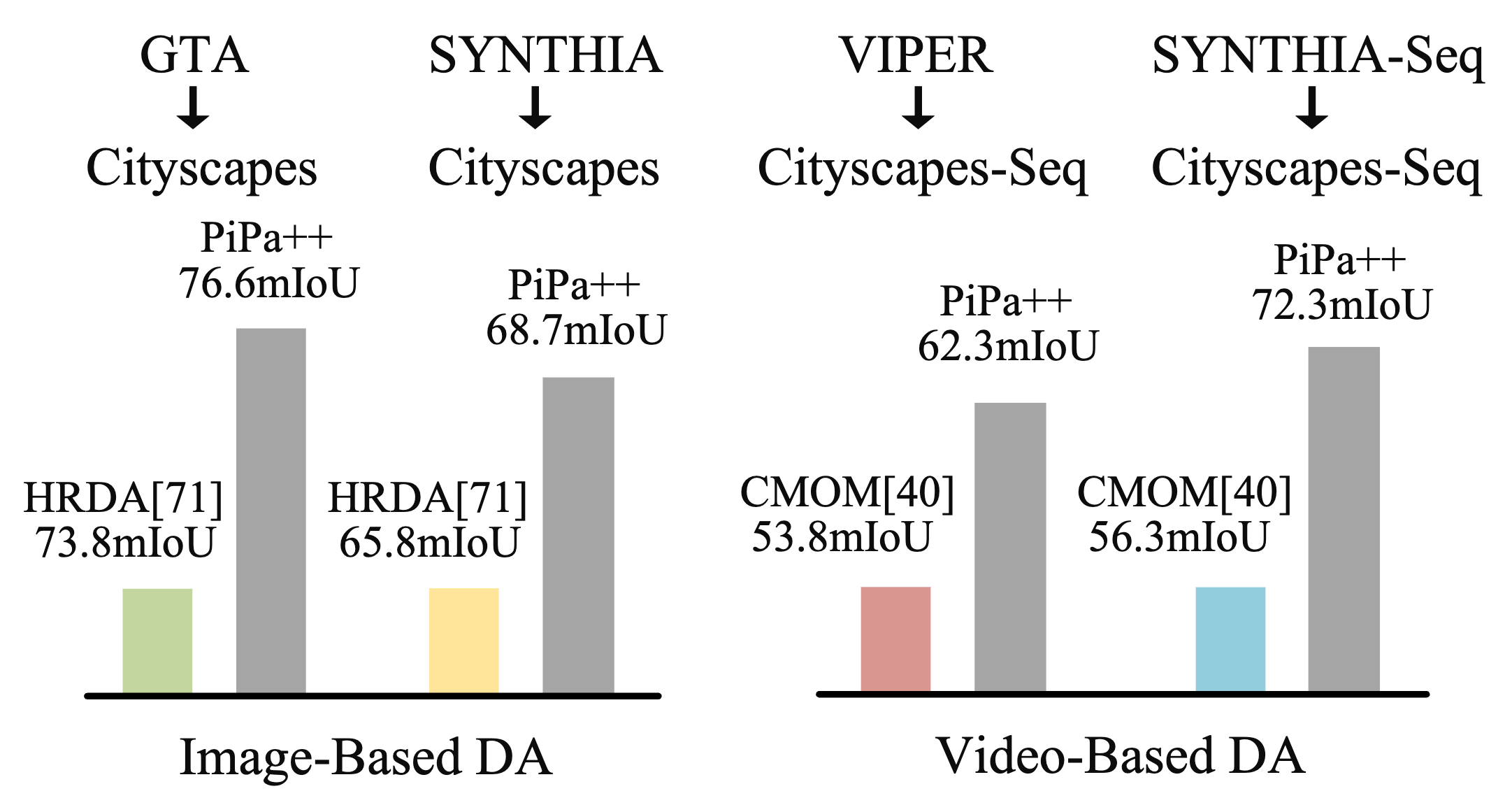}
    \vspace{-0.1in}
    \caption{Different from existing works, we focus on mining the intra-domain knowledge, and argue that the contextual structure between pixels and patches can facilitate the model learning the domain-invariant knowledge in a self-supervised manner. In addition, we exploit the change of contextual structures across in a video and maintain contextual consistency across frames, thereby achieving temporal continuity. Our design embodies efficiency and flexibility, making it perfectly suited for both image-based and video-based domain adaptation tasks, encompassing four competitive datasets: GTA~\cite{richter2016playing} $\rightarrow$ Cityscapes~\cite{MariusCordts2016TheCD}, SYNTHIA~\cite{GermanRos2016TheSD} $\rightarrow$ Cityscapes, SYNTHIA-Seq $\rightarrow$ Cityscapes-Seq, and VIPER~\cite{richter2017playing} $\rightarrow$ Cityscapes-Seq.
    Albeit simple, the proposed learning method is compatible with other existing methods to further boost performance. 
    }
    \label{fig1}
\end{figure}

In recent years, deep neural networks (DNNs) have made significant advancements in the field of computer vision. By processing vast amounts of data, DNNs effectively extract discriminative features and capture subtle patterns from complex visual data, excelling in various visual tasks. This progress has greatly propelled the computer vision applications~\cite{wang2024visual,cheng2023segment,yang2024doraemongpt,chen2024general,chen2024uahoi}, leading to unprecedented achievements in scene understanding tasks such as semantic segmentation, object detection, and scene graph generation. However, deep neural networks are data-hungry and typically require abundant training datasets to be properly trained. For dense prediction tasks such as semantic segmentation, pixel-level annotations are required~\cite{StephanRRichter2016PlayingFD,MariusCordts2016TheCD}. In video scenarios, it is even more demanding~\cite{miao2021vspw}, as dense annotations are needed for each video frame, which entails a prohibitively expensive and time-consuming annotation process. In real-world scenarios, such detailed annotation is hard to meet. To address the shortage of training data, one straightforward idea is to access the abundant synthetic data and the corresponding pixel-level annotations generated by computer graphics. However, there exist domain gaps between synthetic images and real-world images in terms of illumination, weather, and camera parameters. To minimize such a gap, researchers resort to unsupervised domain adaptation (UDA) to transfer the knowledge from labeled source-domain data to the unlabeled target-domain environment.

Prevailing UDA methods are typically designed with highly tailored training paradigms and optimization strategies for static and dynamic scenes. In static scenes, the core research focus is on aligning the source domain and target domain using distribution matching methods such as adversarial training~\cite{hoffman2018cycada, ZuxuanWu2019ACEAT, tsai2018learning,luo2019taking} and self-training~\cite{YangZou2018UnsupervisedDA, zou2019confidence, zheng2021rectifying, yang2020fda}. In dynamic scenes, existing methods~\cite{guan2021domain, wu2022domain, wu2022necessary,xing2023domain} usually leverage optical flow estimation and prediction propagation to bridge distribution shifts across different video domains. While image-level UDA and video-level UDA methods are rapidly improving the state-of-the-art for their benchmarks, they are overwhelmingly task-specific. These methods' specialized designs cannot conceptually generalize to each other, leading to duplicated research and hardware optimization efforts for both tasks. To address this fragmentation, a natural question arises: what must a model possess to effectively tackle both static and dynamic scenarios? Ideally, it should not only 1) attain strong spatial contextual relations but also 2) be well-designed to address temporal continuity at the same time. Prior self-supervised learning methods and their core idea of "learning to compare" have achieved tremendous success in various applications, and they hold significant potential for scene understanding tasks. This success motivates us to rethink the de facto training paradigms of image- and video-semantic segmentation under domain shift. Pixel-level contrast, aside from its basic function of comparing similar classes to distinguish them from other categories to make the networks understand the semantic context better during the training process, should encompass two additional layers of understanding for a dynamic scene to form a good representation space. On one hand, it needs to be aware of the spatial context to prevent totally ignoring the contexts in pixel-level comparisons. On the other hand, humans use ``comparison'' to leverage semantic structures of preceding frames to aid in understanding the current frame. Hence, contrastive learning should enable comparison across frames to achieve temporal continuity and implicit cross-frame correspondences.

With the above insight, we present a novel method, namely PiPa++, which takes advantage of self-supervised learning to naturally achieve a unified architecture by leveraging spatial and temporal representation learning. PiPa++ (1) fully exploits pixel-level contrastive learning to aggregate spatial contextual information and passes the learned spatial information from the source domain to the target domain. (2) Recognizing the rich temporal relationships that can serve as valuable clues for representation learning, it wisely uses the distinguished semantic concepts in historical frames to aid the recognition of the current frame while maintaining temporal continuity.
The choice of contrastive samples is crucial for contrastive learning~\cite{chen2024general,WenguanWang2021ExploringCP}. PiPa++ designs scenario-smart sample mining strategies. For static scenarios, PiPa++ exploits samples from the entire dataset by maintaining a memory bank for each class, thus fully leveraging the dataset's implicit diverse semantic structure to achieve a better representation space. In multi-frame dynamic scenes, PiPa++ restricts the choice of contrastive samples to a limited number of essential frames to prevent temporal inconsistency caused by frames from long distances. To evaluate the performance of our method, we conduct extensive experiments by following previous works on the widely used benchmarks GTA $\rightarrow$ Cityscapes and SYNTHIA $\rightarrow$ Cityscapes for image-level UDA, and SYNTHIA-Seq $\rightarrow$ Cityscapes-Seq, VIPER $\rightarrow$ Cityscapes-Seq for video-level UDA. We achieve notable improvements across these benchmarks (shown in Figure \ref{fig1}). Moreover, our method is compatible with other UDA approaches to further improve performance without introducing extra parameters.

Our main contributions are as follows:

(1) We introduce PiPa++, a unified framework that tackles both image-level and video-level DA under an identical architecture by leveraging self-supervised spatial and temporal representation learning. 

(2) For static environment, different from existing works on inter-domain alignment, we focus on mining domain-invariant knowledge from the original domain
in a self-supervised manner. The proposed Pixel- and Patch-wise self-supervised learning framework harnesses both pixel- and patch-wise consistency against different contexts, which is well-aligned with the segmentation task. For dynamic environment, we maintain temporal continuity without introducing optical flow calculation or additional neural network modules, streamlining the complex pipelines adopted by existing methods.

(3) We propose task-smart sampling strategies to capture more informative samples in both static and dynamic scenarios to aid contrastive learning.

(4) Our self-supervised learning framework does not require extra annotations and is compatible with other existing UDA frameworks. The effectiveness of PiPa++ has been
tested by extensive ablation studies, and it achieves competitive accuracy on four commonly used UDA benchmarks in image- and video- level.

PiPa++ is an extension of our previous conference version, PiPa~\cite{chen2023pipa}. Compared to the previous version, this work introduces the following new content: 

(1) PiPa++ extends the image-level PiPa to a unified architecture that tackles both image-level and video DA tasks without introducing additional parameters. Further experiments are conducted on two widely adopted video-level DA benchmarks.

(2) PiPa++ proposes a task-smart sampling strategy to further boost the performance of PiPa.

(3) To verify the proposed method as a general framework that is orthogonal to existing methods and scalable to different scenarios, we conduct experiments involving multi-source domain adaptation, Clear-to-Adverse-Weather adaptation, and cross-city adaptation. In each case, we observe consistent improvements.

\section{Related Work}
\label{sec: Related}
This work is related to existing work on unsupervised domain adaptation, and contrastive learning.

\paragraph{Unsupervised Domain Adaptation}
Pioneering image-level UDA works \cite{hoffman2018cycada, wu2018dcan} propose to transfer the visual style of the source domain data to the target domain using CycleGAN \cite{zhu2017unpaired}. Later UDA methods can mainly be grouped into two categories according to the technical routes: adversarial training \cite{tsai2018learning, yang2020fda, FeiPan2020UnsupervisedIA, luo2019taking, luo2021category, vu2019advent, wang2021interbn} and self-training \cite{YangZou2018UnsupervisedDA, zou2019confidence, KeMei2020InstanceAS, kang2020adversarial, PanZhang2021PrototypicalPL, WilhelmTranheden2020DACSDA, QianyuZhou2022UncertaintyAwareCR}. Adversarial training methods aim to learn domain-invariant knowledge based on adversarial domain alignment. For instance, Tsai \etal \cite{tsai2018learning} and Luo \etal \cite{luo2019taking} learn domain-invariant representations based on a min-max adversarial optimization game.
However, as shown in \cite{zheng2022adaptive}, unstable adversarial training methods usually lead to suboptimal performance.
Another line of work harnesses self-training to create pseudo labels for the target domain data using the model trained by labeled source domain data. Pseudo labels can be pre-computed either offline \cite{YangZou2018UnsupervisedDA, yang2020fda} or generated online \cite{WilhelmTranheden2020DACSDA, hoyer2022daformer,chen2023pipa,chen2024transferring}. Due to considerable discrepancies in data distributions between two domains, pseudo labels inevitably contains noise. To decrease the influence of faculty labels, Zou \etal \cite{YangZou2018UnsupervisedDA, zou2019confidence} adopts pseudo labels with high confidence. Taking one step further, Zheng \etal \cite{zheng2019unsupervised} conducts the domain alignment to create reliable pseudo labels. Furthermore, some variants leverage specialized sampling \cite{KeMei2020InstanceAS} and uncertainty \cite{zheng2021rectifying} to learn from the noisy  pseudo labels. In addition to the two mainstream practices mentioned above, researchers also conducted extensive attempts such as entropy minimization \cite{vu2019advent, chen2019domain}, image translation \cite{ShaohuaGuo2021LabelFreeRC, JinyuYang2020LabelDrivenRF}, Graph Network \cite{wang2020prototype} and combining adversarial training and self-training \cite{li2019bidirectional, HaoranWang2020ClassesMA,zheng2022adaptive}. Source-free domain adaptation, although a relatively recent concept, has been extensively studied across various fields \cite{you2021domain, li2021imbalanced, ye2022alleviating, huang2022relative}. Recently, Pan \etal \cite{pan2020unsupervised} minimizes the intra-domain discrepancy by separating the target domain into an easy and hard split using an entropy-based ranking function. Yan \etal \cite{yan2021pixel} conducts the inter-domain adaptation between the source and target domain by treating each pixel as an instance. 

Prior video-level UDA methods \cite{guan2021domain, wu2022domain, wu2022necessary} have explored various strategies to tackle domain shifts in video segmentation. Guan \etal \cite{guan2021domain} made the first attempt at video-to-video domain adaptive semantic segmentation, in which both cross-domain and intra-domain temporal consistencies are considered to regularize the learning. Wu \etal \cite{wu2022domain} introduced a novel approach that does not rely on videos for the source domain but instead uses images, leveraging a proxy network to produce pseudo labels for target predictions. Cho \etal \cite{cho2023domain} extended image domain adaptation strategies to video using temporal information by proposing Cross-domain Moving Object Mixing (CMOM) and Feature Alignment with Temporal Context (FATC). Xing \etal \cite{xing2023domain} proposed a method named Temporal Pseudo Supervision (TPS) that explores consistency training in spatiotemporal space for effective domain adaptive video segmentation. 
Different from the above-mentioned works that focus on either image-level or video-level tasks and design-specific architectures, training, and optimization strategies, we utilize the idea of self-supervised learning to design a unified architecture. The proposed method is orthogonal with the above-mentioned approaches, and thus is complementary with existing ones to further boost the result.  

\paragraph{Contrastive Learning}
Contrastive learning is one of the most prominent unsupervised representation learning methods~\cite{AaronvandenOord2018RepresentationLW, ZhirongWu2018UnsupervisedFL, XinleiChen2020ImprovedBW, TingChen2020ASF, KaimingHe2022MomentumCF}, which contrasts similar (positive) data pairs against dissimilar (negative) pairs, thus learning discriminative feature representations.
For instance, Wu \etal~\cite{ZhirongWu2018UnsupervisedFL} learn feature representations at the instance level. He~\etal~\cite{KaimingHe2022MomentumCF} match encoded features to a dynamic dictionary which is updated with a momentum strategy. Chen~\etal \cite{TingChen2020ASF} proposes to engender negative samples from large mini-batches. In the domain adaptative image classification, contrastive learning is utilized to align feature space of different domains \cite{GuoliangKang2019ContrastiveAN, SaeidMotiian2017UnifiedDS}. 
A few recent studies utilize contrastive learning to improve the performance of semantic segmentation task \cite{XinlongWang2020DenseCL, WouterVanGansbeke2021UnsupervisedSS, WenguanWang2021ExploringCP, liu2021domain, BinhuiXie2022SePiCoSP, jiang2022prototypical}. For example, Wang \etal \cite{XinlongWang2020DenseCL} have designed and optimized a self-supervised learning framework for better visual pre-training. Gansbeke \etal \cite{WouterVanGansbeke2021UnsupervisedSS} applies contrastive learning between features from different saliency masks in an unsupervised setting. Recently, Huang \etal \cite{huang2022category} tackles UDA by considering instance contrastive learning as a dictionary look-up operation, allowing learning of category-discriminative feature representations. Xie \etal \cite{xie2021spcl} presents a semantic prototype-based contrastive learning method for fine-grained class alignment. Other works explore contrastive learning either in a fully supervised manner \cite{WenguanWang2021ExploringCP, BinhuiXie2022SePiCoSP} or in a semi-supervised manner \cite{InigoAlonso2021SemiSupervisedSS, zhou2022domain, lai2021semi}. For example, Wang \etal \cite{WenguanWang2021ExploringCP} uses pixel contrast in a fully supervised manner in semantic segmentation. But most methods above either target image-wise instance separation or tend to learn pixel correspondence alone. Different from existing works, we introduce a multi-grained self-supervised learning framework to formulate pixel- and patch-wise contrast in a similar format but at different effect regions. The unified self-supervised learning on both pixel and patch are complementary to each other, and can mine the domain-invariant context feature. Existing works \cite{li2022video,li2023tube,chen2024general} have also attempted to explore temporal knowledge in video segmentation tasks through contrastive learning. However, unlike them, our approach is conducted in an unsupervised manner within the unsupervised domain adaptation (DA) context.

\begin{figure*}
    \centering
    \includegraphics[width=0.9\linewidth]{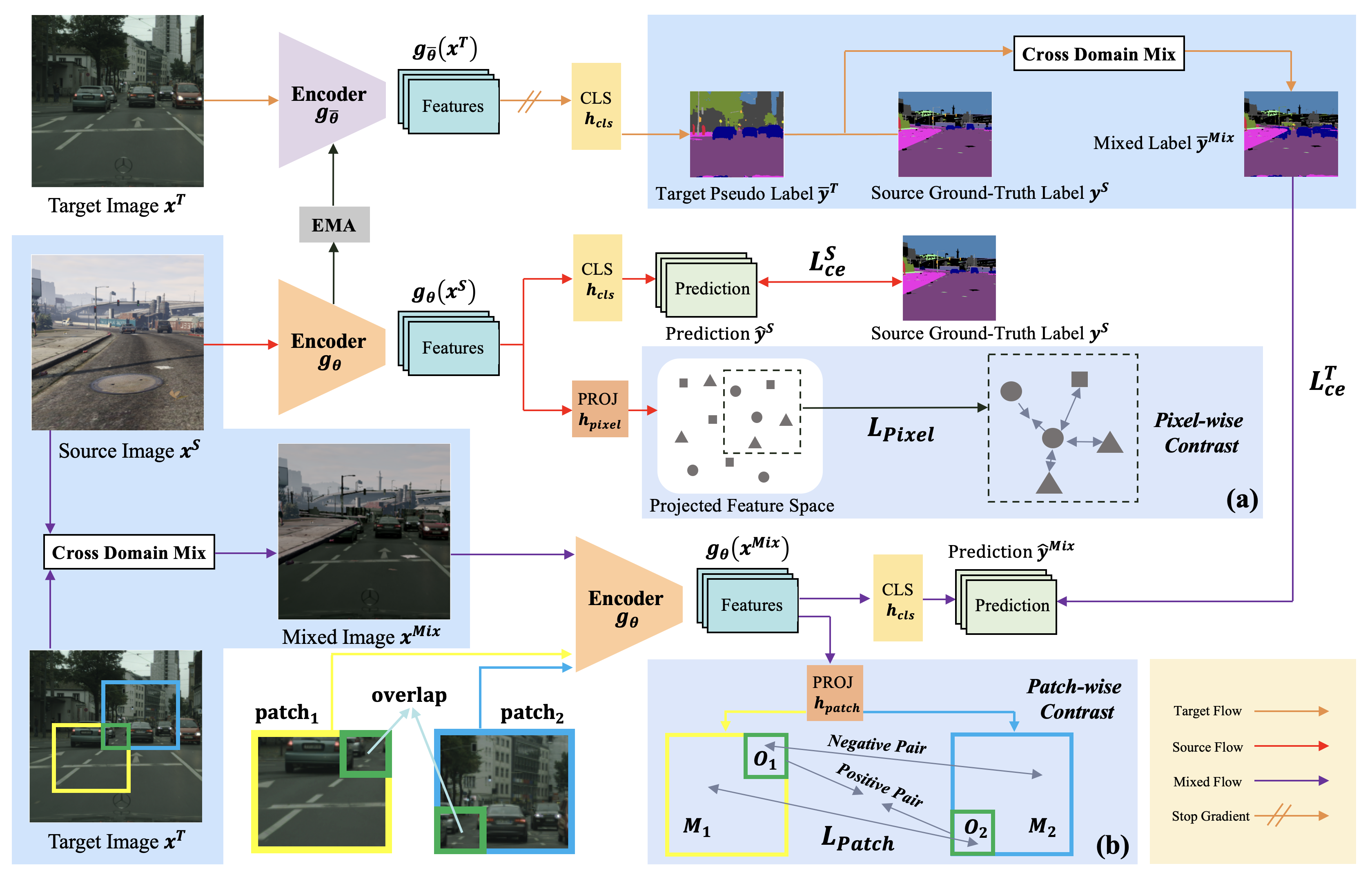}
    \caption{A brief illustration of our unified multi-grained self-supervised learning Framework (PiPa). 
    Given the labeled source data $\left\{\left(x^{S}, y^{S}\right)\right\}$, we calculate the segmentation prediction $\hat{y}^S$ with the backbone $g_\theta$ and the classification head $h_{cls}$, supervised by the basic segmentation loss $L_{ce}^S$.  During training, we leverage the moving averaged model $\boldsymbol{g}_{\bar{\theta}}$ to estimate the pseudo label $\bar{y}^T$ to craft the mixed label $\bar{y}^{Mix}$ based on the category.  According to the mixed label, we copy the corresponding regions as the mixed data $x^{Mix}$. We also deploy the model $g_\theta$ and the head $h_{cls}$ to obtain the mixed prediction $\hat{y}^{Mix}$ supervised by $L_{ce}^T$. 
    Except for the above-mentioned basic segmentation losses, we revisit current pixel contrast and propose a unified multi-grained Contrast. 
    In \textbf{(a)}, we regularize the pixel embedding space by computing pixel-to-pixel contrast: impelling positive-pair embeddings closer, and pushing away the negative embeddings. In \textbf{(b)}, we regularize the patch-wise consistency between projected patch $\mathbf{O}_1$ and $\mathbf{O}_2$. Similarly, we harness the patch-wise contrast, which pulls positive pair, \ie, two features at the same location of $\mathbf{O}_1$ and $\mathbf{O}_2$ closer, while pushing negative pairs apart, \ie, any two features in $\mathbf{M}_1 \cup \mathbf{M}_2$ at different locations. During inference, we drop the two projection heads $h_{patch}$ and $h_{pixel}$ and only keep $g_\theta$ and $h_{cls}$. }
    \label{fig2}
\end{figure*}

\section{Method}
\label{sec: Method}
In this section, we first introduce the problem definition and conventional segmentation losses for semantic segmentation domain adaptation. Then we shed light on the proposed component of our framework PiPa, \ie, Pixel-wise Contrast and Patch-wise Contrast, both of which work on local regions to mine the inherent contextual structures, which is designed for static scenes. Next, we extend PiPa to dynamic scenes to naturally achieve temporal continuity, resulting in PiPa++. We finally also raise a discussion on the mechanism of the proposed method.

\noindent\textbf{Problem Definition.} 
As shown in Figure~\ref{fig2}, given the source-domain synthetic data $X^S=\left\{x^S_u\right\}_{u=1}^U$ labeled by $Y^S=\left\{y_u^{S}\right\}_{u=1}^{U}$ and the unlabelled target-domain real-world data $X^T=\left\{x^T_v\right\}_{v=1}^V$ 
, where $U$ and $V$ are the numbers of images in the source and target domain, respectively. 
The label $Y^S$ belongs to $C$ categories.
Domain adaptive semantic segmentation intends to learn a mapping function that projects the input data $X^T$ to the segmentation prediction $Y^T$ in the target domain.

\noindent\textbf{Basic Segmentation Losses.}  Similar to existing works~\cite{zheng2019unsupervised,zou2019confidence}, we learn the basic source-domain knowledge by adopting the segmentation loss on the source domain 
as:
\begin{equation}
\label{eq:1}
\mathcal{L}_{ce}^S=\mathbb{E}\left[-p_u^S \log h_{cls}(g_\theta\left(x_u^S \right))\right],
\end{equation}
where $p_u^S$ is the 
one-hot vector of the label $y_u^S$, and the value $p_u^S(c)$ equals to 1 if $c==y_u^S$ otherwise 0. 
We harness the visual backbone $g_\theta$, and 2-layer multilayer perceptrons (MLPs) $h_{cls}$ for segmentation category prediction.

To mine the knowledge from the target domain, we generate pseudo labels $\bar{Y}^T = \{\bar{y_v}^T\}$  for the target domain data $X^T$ by a teacher network $g_{\bar{\theta}}$  \cite{WilhelmTranheden2020DACSDA, zhou2022context},
 where $\bar{y_v}^T =  argmax(h_{cls} g_{\bar{\theta}}(x^T_v))$. 
In practice, the teacher network $g_{\bar{\theta}}$ is set as the exponential moving average of the weights of the student network $g_\theta$ after each training iteration \cite{tarvainen2017mean,zheng2022adaptive}. 
Considering that there are no labels for the target-domain data, the network $g_\theta$ is trained on the pseudo label $\bar{y_v}^T$ generated by the teacher model $g_{\bar{\theta}}$. Therefore, the segmentation loss can be formulated as:
\begin{equation}
\label{eq:2}
\mathcal{L}_{ce}^T=\mathbb{E}\left[-\bar{p_v}^T \log h_{cls}( g_\theta\left(x_v^T\right))\right],
\end{equation}
where $\bar{p_v}^T$ is the one-hot vector of the pseudo label $\bar{y_v}^T$.
We observe that pseudo labels inevitably introduce noise considering the data distribution discrepancy between two domains. Therefore, we set a threshold that only the pixels whose prediction confidence is higher than the threshold are accounted for the loss.
In practice, we also follow \cite{WilhelmTranheden2020DACSDA, hoyer2022daformer} to mix images from both domains to facilitate stable training. Specifically, the label $\bar{y}^{Mix}$ is generated by copying the random 50\% categories in $y^S$ and pasting such class areas to the target-domain pseudo label $\bar{y}^T$. Similarly, we also paste the corresponding pixel area in $x^S$ to the target-domain input $x^T$ as $x^{Mix}$.
Therefore, the target-domain segmentation loss is updated as:  
\begin{equation}
\label{eq:3}
\mathcal{L}_{ce}^T=\mathbb{E}\left[-\bar{p_v}^{Mix} \log  h_{cls}(g_\theta\left(x_v^{Mix}\right))\right],
\end{equation}
where $\bar{p_v}^{Mix}$ is the probability vector of the mixed label $\bar{y_v}^{Mix}$. Since we deploy the copy-and-paste strategy instead of the conventional mixup~\cite{zhang2017mixup}, the mixed labels are still one-hot.

\noindent\textbf{Multi-grained Contrast in Different Effect Regions.} 
We note that the above-mentioned segmentation loss does not explicitly consider the inherent context within the image, which is crucial to the local-focused segmentation task. Therefore, we study the feasibility of self-supervised learning in mining intra-domain knowledge for domain adaptive semantic segmentation tasks. In this work, we revisit the current pixel-wise contrast in semantic segmentation \cite{WenguanWang2021ExploringCP} and explore the joint training mechanism of contrastive learning on both pixel- and patch-level effect regions. To this end, we introduce a unified multi-grained contrast including patch-wise contrast to enhance the consistency within a local patch. 

\noindent
In the \textbf{pixel-wise} effect region, given the labels of each pixel $y^S$, we regard image pixels of the same class $C$ as positive samples and the rest pixels in $x^S$ belonging to the other classes are the negative samples. The pixel-wise contrastive loss can be derived as:
\begin{equation}
\label{eq:4}
\mathcal{L}_{\text{Pixel}} =- \sum_{C(i) = C(j)} \log \frac{r\left(e_i, e_j\right)}{\sum_{k=1}^{N_{pixel}} r\left(e_i, e_k\right)},
\end{equation}
where $e$ is the feature map extracted by the projection head $e = h_{pixel}g_{\theta}(x)$, and $N_{pixel}$ is the number of pixels. $e_i$ denotes the $i$-th feature on the feature map $e$.
$r$ denotes the similarity between the two pixel features. In particular, we deploy 
the exponential cosine similarity $r\left(e_i, e_j\right)=\exp \left(s\left(e_i,e_j\right)/ \tau\right)$, where $s$ is cosine similarity between two pixel features $e_i$ and $e_j$, and $\tau$ is the temperature. 
As shown in Figure~\ref{fig2}, with the guide of pixel-wise contrastive loss, the pixel embeddings of the same class are pulled close and those of the other classes are pushed apart, which promotes intra-class compactness and inter-class separability. 
\noindent
In the \textbf{patch-wise} effect region, in particular, given unlabeled target image $x^T$, we also leverage the network $g_\theta$ to extract the feature map of two partially overlapping patches. 
The cropped examples are shown at the bottom of Figure~\ref{fig2}. 
We deploy an independent head $h_{patch}$ with 2-layer MLPs to further project the output feature maps to the embedding space for comparison. 
As shown in Figure \ref{fig2}(b), overlapping region $\mathbf{O}_1$ and $\mathbf{O}_2$ denote the same green area in the original image. In practice, we first randomly select the region $\mathbf{O}$ and then sample two neighbor patches $\mathbf{M}$ covering $\mathbf{O}$. We use $\mathbf{M}$ to denote the entire patch \textbf{including $\mathbf{O}$}.
We argue that the output features of the overlapping region should be invariant to the contexts. Therefore, 
we encourage that each feature in $\mathbf{O}_1$ to be consistent with the corresponding feature of the same location in $\mathbf{O}_2$. Similar to pixel-wise contrast, as shown in Figure~\ref{fig2} module (b), we regard two features at the same position of $\mathbf{O}_1$ and $\mathbf{O}_2$ as positive pair, and any two features in $\mathbf{M}_1$ and $\mathbf{M}_2$ at different positions of the original image are treated as a negative pair. 
Given a target-domain input $x^T$, the patch-wise contrast loss can be formulated as:
\begin{equation}
\label{eq:5}
    \mathcal{L}_{\text {Patch}}=- \sum_{O_1(i)= O_2(j)} \log \frac{r\left({f}_{i}, f_{j}\right)}{\sum_{k=1}^{N_{patch}} r\left({f}_{i}, f_k\right)},
\end{equation}
where $f$ is the feature map extracted by the projection head $f = h_{patch}g_{\theta}(x)$, and $N_{patch}$ is the number of pixels in $M_1 \cup M_2$. $i$ is the pixel index in the patch $\mathbf{M}_1$, and $j$ is for $\mathbf{M}_2$. $\mathbf{O}_1(i)$ denotes the location in the overlapping region $\mathbf{O}_1$. 
$\mathbf{O}_1(i) = \mathbf{O}_2(j)$ denotes $i$ and $j$ are the same pixel (location) in the original image, as shown in Figure \ref{fig2}(b). $f_i$ denotes $i$-th feature in the map. 
Similarly, $r$ denotes the exponential function of the cosine similarity as the one in pixel contrast. 
It is worth noting that we also enlarge the sample pool. In practice, the rest feature $f_k$ not only comes from the union set $M_1 \cup M_2$, but also from other training images within the current batch, which is further explained in task-smart sampling part below.

\begin{figure}[t!]
    \centering
    \includegraphics[width=0.95\linewidth]{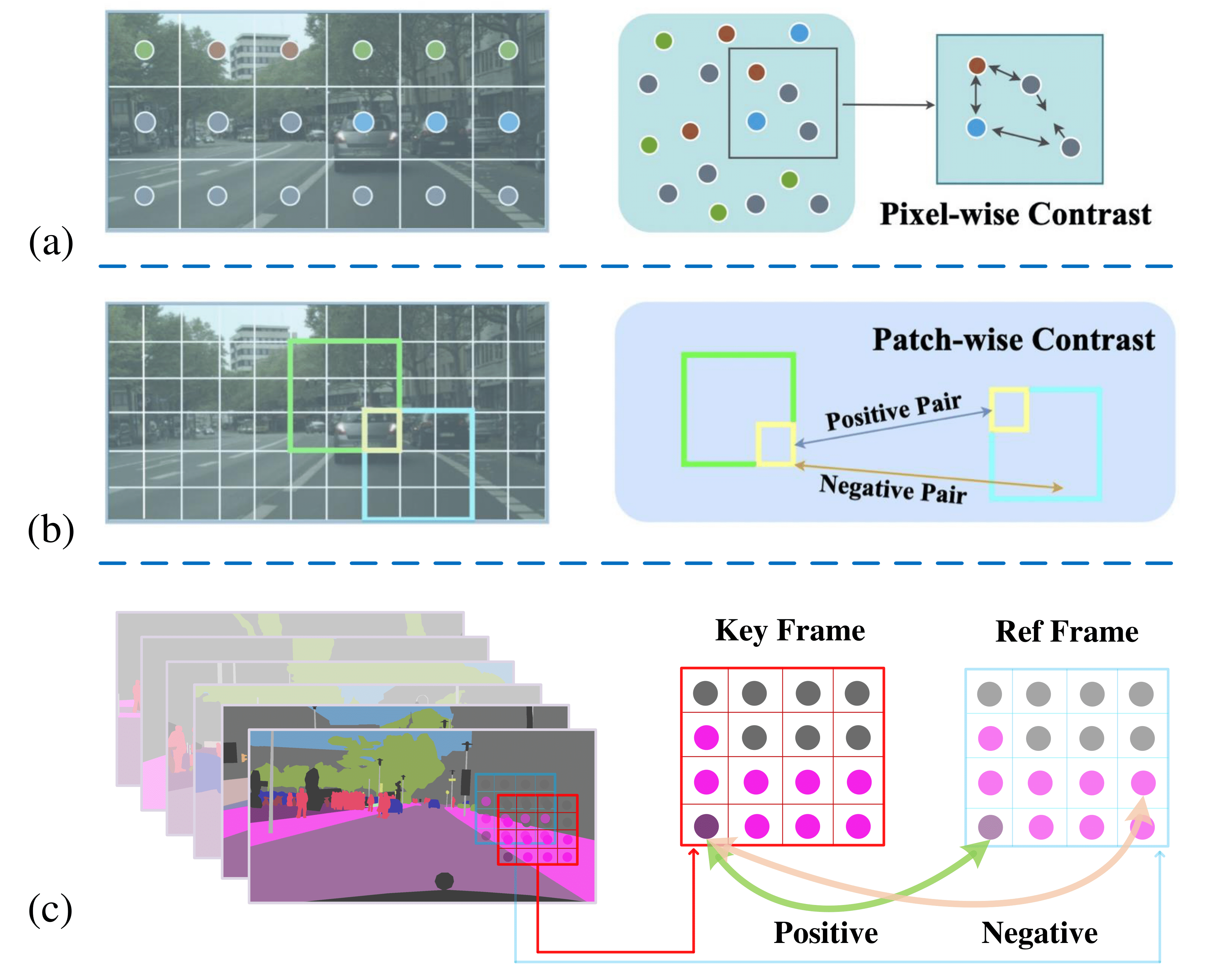}
    \put(-204,5){\small Video Sequence}
    \caption{Our proposed training framework: 
    (a) motivates intra-class compactness and inter-class dispersion by pulling closer the pixel-wise intra-class features and pushing away inter-class features within the image (see a\&b at the top row);
    and  (b) maintains the local patch consistency against different contexts, such as the \textcolor{yellow}{yellow} local patch in the \textcolor{green}{green} and the \textcolor{blue}{blue} patch (see the middle row (b)). (c) aggregates temporal continuity through contrastive leraning across frames.}
    \label{fig7}
\end{figure}

\noindent\textbf{Temporal contrastive learning.} 
Aggregating temporal information is crucial for dynamic scenes. Existing video-level UDA methods typically learn spatial and temporal information in the source domain and transfer them to the target domain. For instance, Cho \etal \cite{cho2023domain} utilized moving objects and their temporal information from the source domain, mixing them into the background of the target domain. This allows the model to observe the motion of source domain objects within the target domain's background, thereby learning better spatiotemporal features. This cross-domain mixing strategy helps enhance the model's temporal continuity and segmentation performance in the target domain. However, this method requires task-specific design, limiting its flexibility. Other works enhance temporal continuity in the target domain by utilizing motion information between video frames, but the introduction of optical modules incurs significant computational overhead.
To simplify the cumbersome pipelines in domain adaptive video semantic segmentation and achieve a unified framework, we extend PiPa \cite{chen2023pipa} to PiPa++, whose core idea is to extend PiPa's capability of capturing context information through multi-grained contrastive learning. PiPa++ leavrages cross-frame temporal contrast to learn temporal continuity in the target domain.
As shown in Figure \ref{fig7}(c), when selecting a key frame $f_{key}$ during training, a reference frame from its temporal neighborhood is also randomly selected, which is denoted as $f_{ref}$. Typically, in two adjacent or closely spaced frames, the temporal information remains consistent, as evidented by \cite{li2023tube,gan2016you,wu2022seqformer,li2022video}. Following similar idea of patch-wise contrastive learning, we crop adjacent frames $f_{key}$ and $f_{ref}$ to two overlapping patches. Then we adopt another lightweight projection head $h_{temp}$ with 2-layer MLPs to project the output feature maps of the two adjacent frames to the embedding space for comparison. The cropped patches are $ft_{key}$ and $ft_{ref}$ after projection. In video scenario, a pixel at $ft_{ref}$ is defined as positive if it is at the same location at $ft_{key}$ and all other pixels inside $ft_{ref}$ are considered negative. To maintain the contextual structure between to frames and hence, realizes temporal continuity, the positive samples are pulled together and negative samples are pushed away:

\begin{equation}
\label{eq:temporal_loss}
\mathcal{L}_{\text{Temp}} = -\sum_{i = j} \log \frac{r(f_{key_i}, f_{ref_j})}{\sum_{k=1}^{N_{temp}} r(f_{key_i}, f_{ref_k})},
\end{equation}
where $f_{key_i}$, $f_{ref_j}$ and $f_{ref_k}$ are pixel features extracted by the temporal projection head $h_{temp}$, denoting key pixel sample, reference positive and reference negative samples, respectively. $N_{temp}$ is the number of pixels in $ft_{key} \cup ft_{ref}$. $i$ is the pixel index in the key frame patch, and $j$ is for the reference frame patch. $i = j$ denotes that $i$ and $j$ are the same pixel (location) in the original image. $f_i$ denotes the $i$-th feature in the map. Similarly, $r$ denotes the exponential function of the cosine similarity, as used in patch-wise contrast. 

\noindent\textbf{Task-smart Sampling.} 
Prior works has validated that the choice of contrastive sampling significantly enhances the discriminating power of representation space \cite{kalantidis2020hard,WenguanWang2021ExploringCP,chen2024general}. In our case, informative samples differ in static and dynamic scenes, leading to the formation of a task-smart sampling strategy.
In static scenes, to obtain a large number of highly structured pixel training samples, we utilize a semantic-aware memory bank to store not just training samples from a single image, but rich semantic information extracted from the same batch, and even the whole dataset. This design allows us to access more representative data samples obtained from the entire dataset during each training step.
For dynamic scenes, the data usually comes in the form of long video sequences, and temporal continuity typically remains consistent only between adjacent frames. Over time, the objects in the scene and their contextual structures usually undergo significant changes. Including such data in our contrastive framework would lead to performance degradation. Therefore, we select positive and negative samples only from a range of 1 to 3 frames. Further experimental validation are provided in the ablation study section. 

\noindent\textbf{Total Loss.} The overall training objective is the combination of pixel-level cross-entropy loss and the proposed PiPa++: 
\begin{equation}
\label{eq:6}
\mathcal{L}_{total}=\mathcal{L}_{\mathrm{ce}}^S+\mathcal{L}_{\mathrm{ce}}^T+\alpha \mathcal{L}_{\text {Pixel}}+\beta \mathcal{L}_{\text {Patch}} + \gamma \mathcal{L}_{\text{Temp}},
\end{equation}
where 
$\alpha$, $\beta$, and $\gamma$ are the weights for pixel-wise contrast $\mathcal{L}_{\text{Pixel}}$, patch-wise contrast $\mathcal{L}_{\text{Patch}}$, and temporal contrast $\mathcal{L}_{\text{Temp}}$ respectively. Temporal contrast $\mathcal{L}_{\text{Temp}}$ is not applied at static scene.

\noindent\textbf{Discussion.} 
\noindent\textbf{1. Correlation between Pixel and Patch Contrast.} Both pixel and patch contrast are derived from instance-level contrastive learning and share a common underlying idea, \ie, contrast, but they work at different effect regions, \ie, pixel-wise and patch-wise. 
The pixel contrast explores the pixel-to-pixel category correlation over the whole image, while patch-wise contrast imposes regularization on the semantic patches from a local perspective. 
Therefore, the two kinds of contrast are complementary and can work in a unified way to mine the intra-domain inherent context within the data.
\noindent\textbf{2. What is the advantage of the proposed framework?} Traditional UDA methods focus on learning shared inter-domain knowledge. Differently, we are motivated by the objectives of UDA semantic segmentation in a bottom-up manner, and thus leverage rich pixel correlations in the training data to facilitate intra-domain knowledge learning. 
By explicitly regularizing the feature space via PiPa, we enable the model to explore the inherent intra-domain context in a self-supervised setting, \ie, pixel-wise and patch-wise,  without extra parameters or annotations. 
Therefore, PiPa could be effortlessly incorporated into existing UDA approaches to achieve better results without extra overhead during testing.
\noindent\textbf{3. Difference from conventional contrastive learning.}
Conventional contrastive learning methods typically tend to perform contrast in the instance or pixel level alone \cite{ZhirongWu2018UnsupervisedFL, huang2022category, WenguanWang2021ExploringCP}. 
We formulate pixel- and patch-wise contrast in a similar format but focus on the local effect regions within the images, which is well aligned with the local-focused segmentation task. 
We show that the proposed local contrast, \ie, pixel- and patch-wise contrasts, regularizes the domain adaptation training and guides the model to shed more light on the intra-domain context. 
Our experiment also verifies this point that pixel- and patch-wise contrast facilitates smooth edges between different categories and yields a higher accuracy on small objects.
\noindent\textbf{4. Why PiPa++ does not perform temporal contrastive learning on source domain.}
In general, for static scenes, the contextual information learned from the source domain needs to be passed to the target domain to achieve good performance in the target domain. However, for temporal information, PiPa++ focuses more on learning cross-frame continuity. This is mainly achieved through comparison between key and reference frames. PiPa++ learns temporal information solely within the target domain's video sequences without the need to introduce additional networks such as optical flow. Thus, it achieves a unified network that simultaneously handles both image and video domain adaptation.

\subsection{Implementation Details} 
\noindent\textbf{Structure Details.}
Following recent SOTA UDA setting \cite{hoyer2022daformer,zhou2022context, BinhuiXie2022SePiCoSP}, our network consists of a SegFormer MiT-B5 backbone~\cite{xie2021segformer,hoyer2022daformer} pretrained on ImageNet-1k~\cite{deng2009imagenet} and several MLP-based heads, \ie, $h_{\text {cls }}$, $h_{\text {pixel }}$ and $h_{\text {patch }}$, which contains two fully-connected (fc) layers and ReLU activation between two fc layers. 
Note that the self-supervised projection heads $h_{\text {pixel }}$ and $h_{\text {patch }}$ are only applied at training time and are removed during inference, which does not introduce extra computational costs in deployment. 

\noindent\textbf{Implementation details.}
We train the network with batch size 2 for 60k iterations with a single NVIDIA RTX 6000 GPU. 
We adopt AdamW \cite{loshchilov2017decoupled} as the optimizer, a learning rate of $6 \times 10^{-5}$, a linear learning rate warmup of $1.5 \mathrm{k}$ iterations and the weight decay of 0.01. 
Following \cite{zhou2022context, BinhuiXie2022SePiCoSP}, the input image is resized to 1280 $\times$ 720 for GTA and 1280 $\times$ 760 for SYNTHIA, with a random crop size of 640 $\times$ 640. For the patch-wise contrast, we randomly resize the input images by a ratio between 0.5 and 2, and then randomly crop two patches of the size $720 \times 720$ from the resized image and ensure the Intersection-over-Union(IoU) value of the two patches between 0.1 and 1. 
We utilize the same data augmentation \eg, color jitter, Gaussian blur and ClassMix \cite{olsson2021classmix} and empirically set pseudo labels threshold $0.968$ following \cite{WilhelmTranheden2020DACSDA}. The exponential moving average parameter of the teacher network is $0.999$. The hyperparameters of the loss function are chosen empirically $\alpha = \beta = \gamma = 0.1$.

\section{Experiments}
\label{sec: Experiment}

\subsection{Results for Image-baesd UDA}
\noindent\textbf{Datasets.} 
We evaluate the proposed method on image-based UDA benchmarks GTA $\rightarrow$ Cityscapes and SYNTHIA $\rightarrow$ Cityscapes, following common UDA protocols \cite{WilhelmTranheden2020DACSDA, wang2021domain, araslanov2021self, hoyer2022daformer, hoyer2022hrda}. The target dataset, Cityscapes, is collected from real-world street-view images and consists of 2,975 unlabeled images for training, 500 images for validation, and 1,525 images for testing. Cityscapes provides high-quality pixel-level annotations for urban street scenes, making it a challenging and realistic benchmark for evaluating domain adaptation methods. The GTA dataset is a synthetic dataset generated from the video game Grand Theft Auto V, containing 24,966 images with high-quality annotations for 19 classes. It is widely used in domain adaptation research due to its diverse and complex urban environments. Similarly, the SYNTHIA dataset is another synthetic dataset that includes 9,400 images with pixel-level annotations for 13 classes. It provides a variety of scenes and weather conditions, enhancing the robustness of domain adaptation models. We report the results on the Cityscapes validation set for comparisons, as it is a standard practice in the community.

\noindent\textbf{Performance.}
Generally, our PiPa yields a significant improvement over the transformer-based models DAFormer\cite{hoyer2022daformer} and HRDA\cite{hoyer2022hrda}. Particularly, PiPa achieves 71.7 mIoU, which outperforms DAFormer by a considerable margin of +3.4 mIoU. Additionally, when applying PiPa to HRDA, which is a strong baseline that adopts high-resolution crops, we increase +1.8 mIoU and achieve the state-of-the-art performance of 75.6 mIoU, verifying the effectiveness of the proposed method that introduces a unified and multi-grained self-supervised learning algorithm in UDA task. Furthermore, PiPa achieves leading IoU of almost all classes on GTA $\rightarrow$ Cityscapes, including several small-scale objectives such as Fence, Pole, Wall and Training Sign. Particularly, we increase the IoU of the Fence by +6.2 from 51.5 to 57.7 IoU. The IoU performance of PiPa verifies our motivation that the exploration of the inherent structures of intra-domain images indeed helps category recognition, especially for challenging small objectives. As revealed in Table~\ref{table:syncity}, PiPa also achieves remarkable mIoU and mIoU* (13 most common categories) performance on SYNTHIA $\rightarrow$ Cityscapes, increasing +2.5 and +2.4 mIoU compared with DAFormer \cite{hoyer2022daformer} and HRDA \cite{hoyer2022hrda}, respectively.

\begin{figure*}[!t]
    \centering
    \includegraphics[width=0.95\linewidth]{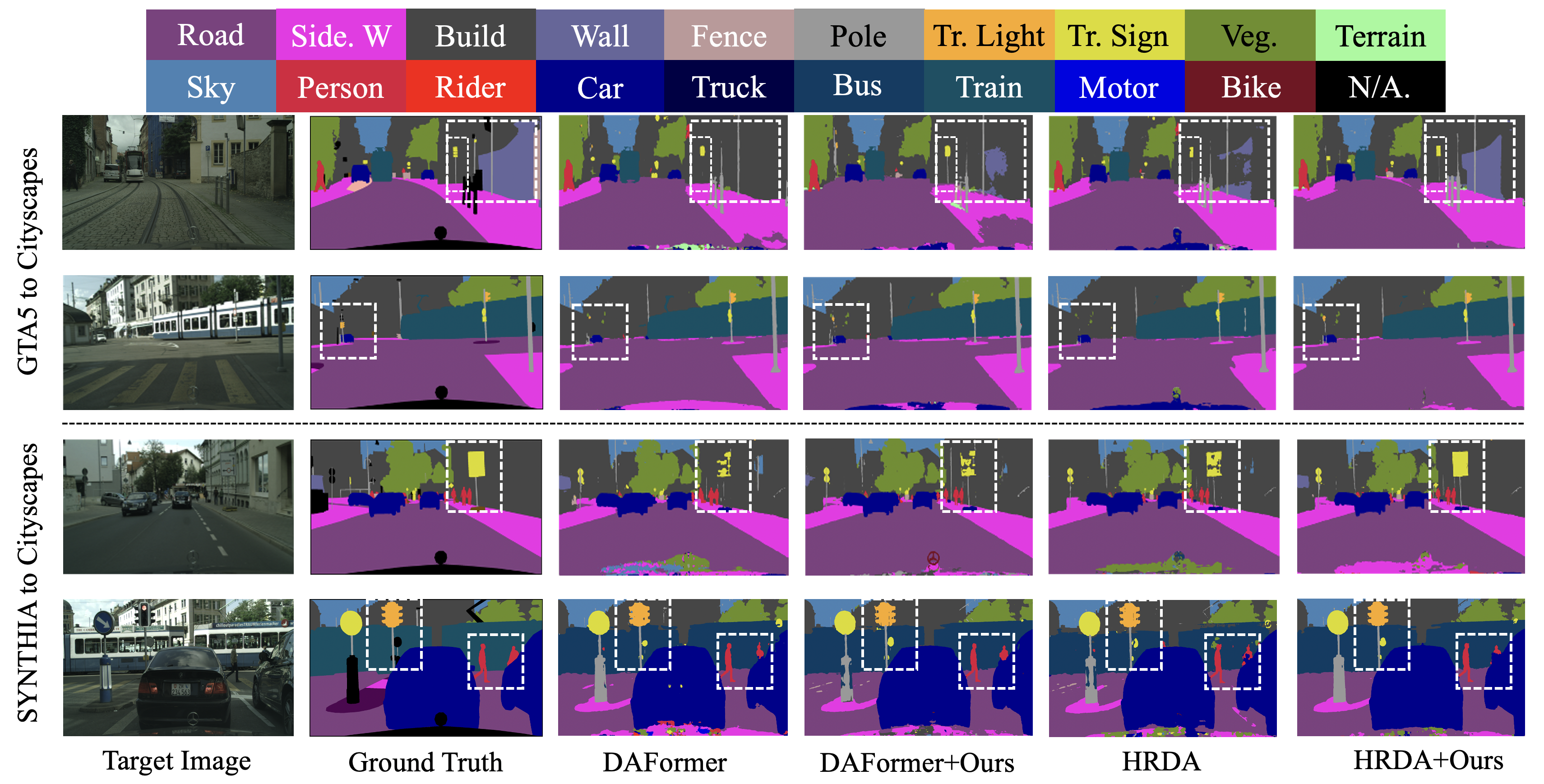}
    \caption{Qualitative results on GTA $\rightarrow$ Cityscapes and SYNTHIA $\rightarrow$ Cityscapes. From left to right: Target Image, Ground Truth, the visual results predicted by DAFormer, DAFormer + Ours (PiPa), HRDA, HRDA + Ours (PiPa). We deploy the white dash boxes to highlight different prediction parts. }
    \label{fig3}
\end{figure*}

\noindent\textbf{Qualitative results.}
In Figure~\ref{fig3}, we visualize the segmentation results and the comparison with previous strong methods DAFormer \cite{hoyer2022daformer}, HRDA \cite{hoyer2022hrda}, and the ground truth on both GTA $\rightarrow$ Cityscapes and SYNTHIA $\rightarrow$ Cityscapes benchmarks. The results highlighted by white dash boxes show that PiPa is capable of segmenting minor categories such as `wall', `traffic sign' and `traffic light'. It is also noticeable that PiPa predicts smoother edges between different categories, \eg, `person' in the fourth row of Figure~\ref{fig3}. We think it is because the proposed method explicitly encourages patch-wise consistency against different contexts, which facilitates the prediction robustness on edges.

\subsection{Results for Video-baesd UDA}
\noindent\textbf{Datasets.} 
We evaluate the proposed method on video-based UDA benchmarks VIPER $\rightarrow$ Cityscapes-Seq and SYNTHIA-Seq $\rightarrow$ Cityscapes-Seq, following common UDA protocols. The target dataset, Cityscapes-Seq, is an extension of the Cityscapes dataset, providing sequences of images captured in real-world urban environments. It consists of video sequences with pixel-level annotations, offering a challenging and realistic benchmark for evaluating domain adaptation methods in video segmentation tasks. The VIPER dataset is a synthetic video dataset generated from the game Grand Theft Auto V, containing 254,064 frames with high-quality annotations for 32 classes. It provides diverse and complex urban scenarios, making it a valuable resource for domain adaptation research. Similarly, the SYNTHIA-Seq dataset is another synthetic video dataset that includes sequences of images with pixel-level annotations for 13 classes. It offers various scenes and weather conditions, enhancing the robustness of domain adaptation models. We report the results on the Cityscapes-Seq validation set for comparisons, as it is a standard practice in the community.

\noindent\textbf{Performance.}
We conducted extensive experiments to evaluate the performance of our proposed method PiPa++ on the video-based UDA tasks. The quantitative results on the VIPER $\rightarrow$ Cityscapes-Seq and SYNTHIA-Seq $\rightarrow$ Cityscapes-Seq benchmarks are presented in Tables \ref{table_viper} and \ref{table_video_syn}, respectively. Our method achieves significant improvements over the baseline methods. Specifically, on the VIPER $\rightarrow$ Cityscapes-Seq task, PiPa++ outperforms the state-of-the-art method CMOM by 8.5 mIoU, reaching a new state-of-the-art performance of 62.3 mIoU. On the SYNTHIA-Seq $\rightarrow$ Cityscapes-Seq task, PiPa++ achieves an impressive 72.3 mIoU, which is 16.0 mIoU higher than the previous best method. These results demonstrate the effectiveness of our unified framework in leveraging both spatial and temporal information to improve segmentation performance in video-based UDA tasks.

\begin{figure*}[!t]
    \centering
    \includegraphics[width=0.95\linewidth]{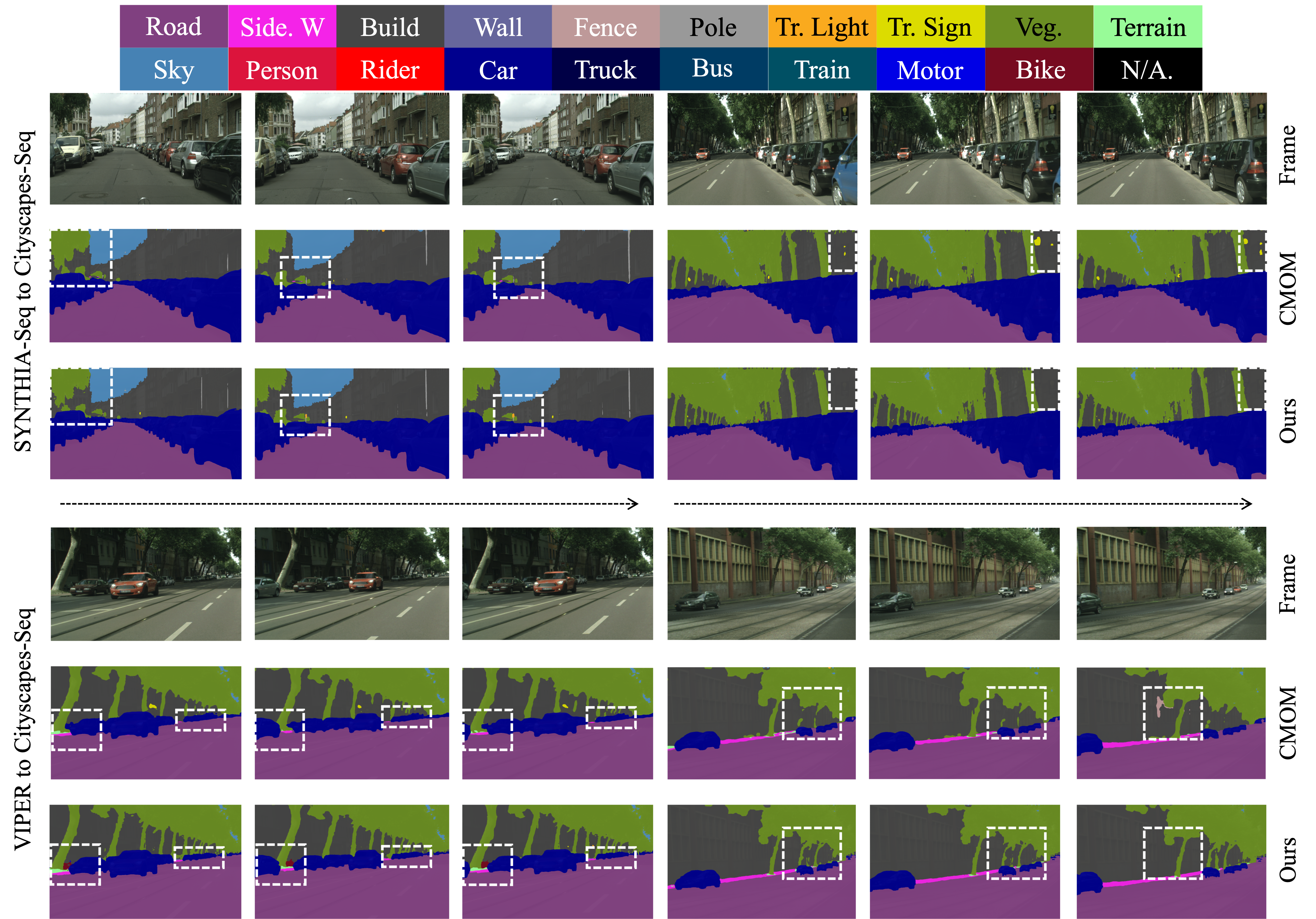}
    \caption{Qualitative results on VIPER $\rightarrow$ Cityscapes-Seq and SYNTHIA-Seq $\rightarrow$ Cityscapes-Seq. From left to right: Target Image, Ground Truth, the visual results predicted by DAFormer, DAFormer + Ours (PiPa), HRDA, HRDA + Ours (PiPa). We deploy the white dash boxes to highlight different prediction parts. }
    \label{fig5}
\end{figure*}

\noindent\textbf{Qualitative results.}
To further validate the effectiveness of our method, we provide qualitative comparisons of segmentation results on VIPER $\rightarrow$ Cityscapes-Seq and SYNTHIA-Seq $\rightarrow$ Cityscapes-Seq in Figure \ref{fig5}. The visual results clearly show that our proposed method PiPa++ can produce more accurate and consistent segmentation maps compared to the baseline methods. For instance, PiPa++ successfully segments small and challenging objects like poles and traffic lights, which are often missed or misclassified by other methods. Additionally, our method maintains better temporal consistency across frames, as evidenced by the smoother and more coherent segmentation results. The qualitative improvements, along with the quantitative gains, highlight the robustness and generalization capabilities of PiPa++ in handling video-based domain adaptation tasks.

\begin{table*}[!t]
    \centering
    \caption{Quantitative comparison with previous image-based UDA methods on GTA $\rightarrow$ Cityscapes. We present pre-class IoU and mIoU. The best accuracy in every column is in \textbf{bold}.}
    \label{table:gtacity}
    \resizebox{\linewidth}{!}{
    \begin{tabular}{c|ccccccccccccccccccc|c}
        \shline
        Method & Road & SW & Build & Wall & Fence & Pole & TL & TS & Veg. & Terrain & Sky & PR & Rider & Car & Truck & Bus & Train & Motor & Bike & mIoU\\
        \shline
        CyCADA \cite{hoffman2018cycada} & 86.7 & 35.6 & 80.1 & 19.8 & 17.5 & 38.0 & 39.9 & 41.5 & 82.7 & 27.9 & 73.6 & 64.9 & 19.0 & 65.0 & 12.0 & 28.6 & 4.5 & 31.1 & 42.0 & 42.7 \\
        CLAN~\cite{luo2019taking} & 87.0 & 27.1 & 79.6 & 27.3 & 23.3 & 28.3 & 35.5 & 24.2 & 83.6 & 27.4 & 74.2 & 58.6 & 28.0 & 76.2 & 33.1 & 36.7 & 6.7 & 31.9 & 31.4 & 43.2 \\
        ASA~\cite{zhou2020affinity} & 89.2 & 27.8 & 81.3 & 25.3 & 22.7 & 28.7 & 36.5 & 19.6 & 83.8 & 31.4 & 77.1 & 59.2 & 29.8 & 84.3 & 33.2 & 45.6 & 16.9 & 34.5 & 30.8 & 45.1 \\
        SPCL \cite{xie2021spcl} & 90.3 & 50.3 & 85.7 & 45.3 & 28.4 & 36.8 & 42.2 & 22.3 & 85.1 & 43.6 & 87.2 & 62.8 & 39.0 & 87.8 & 41.3 & 53.9 & 17.7 & 35.9 & 33.8 & 52.1\\
        DACS \cite{WilhelmTranheden2020DACSDA} & 89.9 & 39.7 & 87.9 & 30.7 & 39.5 & 38.5 & 46.4 & 52.8 & 88.0 & 44.0 & 88.8 & 67.2 & 35.8 & 84.5 & 45.7 & 50.2 & 0.0 & 27.3 & 34.0 & 52.1\\
        BAPA \cite{liu2021bapa} & 94.4 & 61.0 & 88.0 & 26.8 & 39.9 & 38.3 & 46.1 & 55.3 & 87.8 & 46.1 & 89.4 & 68.8 & 40.0 & 90.2 & 60.4 & 59.0 & 0.0 & 45.1 & 54.2 & 57.4\\
        CaCo \cite{huang2022category} & 93.8 & 64.1 & 85.7 & 43.7 & 42.2 & 46.1 & 50.1 & 54.0 & 88.7 & 47.0 & 86.5 & 68.1 & 2.9 & 88.0 & 43.4 & 60.1 & 31.5 & 46.1 & 60.9 & 58.0\\
        PiPa (CNN) & 95.1 & 71.3 & 87.7 & 44.2 & 42.0 & 43.5 & 52.1 & 63.3 & 87.8 & 44.0 & 87.5 & 72.3 & 44.2 & 89.3 & 59.9 & 59.4 & 2.1 & 47.2 & 48.9 & 60.1 \\
        \hline
        DAFormer \cite{hoyer2022daformer} & 95.7 & 70.2 & 89.4 & 53.5 & 48.1 & 49.6 & 55.8 & 59.4 & 89.9 & 47.9 & 92.5 & 72.2 & 44.7 & 92.3 & 74.5 & 78.2 & 65.1 & 55.9 & 61.8 & 68.3\\
        CAMix \cite{zhou2022context} & 96.0 & 73.1 & 89.5 & 53.9 & 50.8 & 51.7 & 58.7 & 64.9 & 90.0 & 51.2 & 92.2 & 71.8 & 44.0 & 92.8 & 78.7 & 82.3 & 70.9 & 54.1 & 64.3 & 70.0\\
        \rowcolor{lightgray} 
        DAFormer \cite{hoyer2022daformer} + PiPa  & 96.1 & 72.0 & 90.3 & 56.6 & 52.0 & 55.1 & 61.8 & 63.7 & 90.8 & 52.6 & 93.6 & 74.3 & 43.6 & 93.5 & 78.4 & 84.2 & 77.3 & 59.9 & 66.7 & 71.7\\
        \hline
        HRDA \cite{hoyer2022hrda} & 96.4 & 74.4 & 91.0 & 61.6 & 51.5 & 57.1 & 63.9 & 69.3 & 91.3 & 48.4 & 94.2 & 79.0 & 52.9 & 93.9 & 84.1 & 85.7 & 75.9 & 63.9 & 67.5 & 73.8\\
        \rowcolor{lightgray} 
        HRDA \cite{hoyer2022hrda} + PiPa & 96.8 & 76.3 & 91.6 & \textbf{63.0} & 57.7 & 60.0 & 65.4 & 72.6 & \textbf{91.7} & 51.8 & \textbf{94.8} & 79.7 & 56.4 & 94.4 & 85.9 & 88.4 & 78.9 & 63.5 & 67.2 & 75.6\\
        \rowcolor{lightgray}
        HRDA \cite{hoyer2022hrda} + PiPa++ & 96.4 & \textbf{77.1} & \textbf{92.1} & 62.5 & \textbf{58.4} & \textbf{63.1} & \textbf{67.3} & \textbf{75.9} & 91.0 & \textbf{53.4} & 94.2 & \textbf{83.5} & \textbf{59.1} & \textbf{95.1} & \textbf{86.0} & \textbf{89.8} & \textbf{81.9} & 62.9 & 66.5 & \textbf{76.6}\\
        \shline
        \shline
    \end{tabular}
    }
\end{table*}

\begin{table*}[!t]
    \centering
    \caption{Quantitative comparison with previous UDA methods on SYNTHIA $\rightarrow$ Cityscapes. We present pre-class IoU, mIoU and mIoU*. mIoU and mIoU* are averaged over 16 and 13 categories, respectively. The best accuracy in every column is in \textbf{bold}.}
    \label{table:syncity}
    \resizebox{\linewidth}{!}{
    \begin{tabular}{c|cccccccccccccccc|c|c}
        \shline
        Method & Road & SW & Build & Wall* & Fence* & Pole* & TL & TS & Veg. & Sky & PR & Rider & Car & Bus & Motor & Bike & mIoU* & mIoU\\
        \shline
        CLAN~\cite{luo2019taking} & 81.3 & 37.0 & 80.1 & $-$ & $-$ & $-$ & 16.1 & 13.7 & 78.2 & 81.5 & 53.4 & 21.2 & 73.0 & 32.9 & 22.6 & 30.7 & 47.8 & $-$  \\
        SP-Adv~\cite{SHAN2020125} & 84.8 & 	35.8 & 	78.6 & $-$ & $-$ & $-$ &	6.2	& 15.6 & 	80.5 &	82.0 &	66.5 & 	22.7 & 	74.3 &	34.1	 & 19.2	 & 27.3 & 	48.3 &  $-$ \\
        ASA~\cite{zhou2020affinity} & \textbf{91.2} & 48.5 & 80.4 & 3.7 & 0.3 & 21.7 & 5.5 & 5.2 & 79.5 & 83.6 & 56.4 & 21.0 & 80.3 & 36.2 & 20.0 & 32.9 & 49.3 & 41.7 \\
        DADA~\cite{vu2019dada} & 89.2 & 44.8 & 81.4 & 6.8 & 0.3 & 26.2 & 8.6 & 11.1 & 81.8 & 84.0 & 54.7 & 19.3 & 79.7 & 40.7 & 14.0 & 38.8 & 49.8 & 42.6 \\
        CCM~\cite{li2020content} & 79.6 & 36.4 & 80.6 & 13.3 & 0.3 & 25.5 & 22.4 & 14.9 & 81.8 & 77.4 & 56.8 & 25.9 & 80.7 & 45.3 & 29.9 &  52.0 & 52.9 & 45.2 \\
        BL~\cite{li2019bidirectional} & 86.0 & 46.7 & 80.3 & $-$ & $-$ & $-$ & 14.1 & 11.6 & 79.2 & 81.3 & 54.1 & 27.9 & 73.7 & 42.2 & 25.7 & 45.3 & 51.4 & $-$ \\
        \hline
        DAFormer \cite{hoyer2022daformer} & 84.5 & 40.7 & 88.4 & 41.5 & 6.5 & 50.0 & 55.0 & 54.6 & 86.0 & 89.8 & 73.2 & 48.2 & 87.2 & 53.2 & 53.9 & 61.7 & 67.4 & 60.9 \\
        CAMix \cite{zhou2022context} & 87.4 & 47.5 & 88.8 & $-$ & $-$ & $-$ & 55.2 & 55.4 & 87.0 & 91.7 & 72.0 & 49.3 & 86.9 & 57.0 & 57.5 & 63.6 & 69.2 & $-$ \\
        \rowcolor{lightgray} 
        DAFormer \cite{hoyer2022daformer} + PiPa & 87.9 & 48.9 & 88.7 & 45.1 & 4.5 & 53.1 & 59.1 & 58.8 & 87.8 & 92.2 & 75.7 & 49.6 & 88.8 & 53.5 & 58.0 & 62.8 & 70.1 & 63.4 \\
        \hline
        HRDA \cite{hoyer2022hrda} & 85.2 & 47.7 & 88.8 & 49.5 & 4.8 & 57.2 & 65.7 & 60.9 & 85.3 & 92.9 & 79.4 & 52.8 & 89.0 & 64.7 & 63.9 & 64.9 & 72.4 & 65.8 \\
        \rowcolor{lightgray} 
        HRDA \cite{hoyer2022hrda} + PiPa & 88.6 & \textbf{50.1} & 90.0 & \textbf{53.8} & 7.7 & 58.1 & 67.2 & 63.1 & 88.5 & 94.5 & \textbf{79.7} & 57.6 & \textbf{90.8} & 70.2 & 65.1 & 66.9 & 74.8 & 68.2 \\
        \rowcolor{lightgray} 
        HRDA \cite{hoyer2022hrda} + PiPa++ & \textbf{89.4} & 49.6 & \textbf{90.9} & 53.3 & \textbf{8.5} & \textbf{58.7} & \textbf{68.1} & \textbf{64.4} & \textbf{89.0} & \textbf{95.1} & 79.0 & \textbf{58.4} & 90.1 & \textbf{70.9} & \textbf{66.2} & \textbf{67.8} & \textbf{75.3} & \textbf{68.7} \\
        \shline
    \end{tabular}
    }
\end{table*}

\begin{table*}[!ht]
    \caption{Quantitative comparison with previous video-based UDA methods on VIPER~$\rightarrow$~Cityscapes-Seq. We present per-class IoU and mIoU. The best accuracy in every column is in bold.}
    \resizebox{\linewidth}{!}{
    \centering
    \begin{tabular}{c|c|ccccccccccccccccccc|c}
     \shline
     Method &Training &{Road} &{SW} &{Build} &{Fence} &{TL} &{TS} &{Veg.} &{Terrain} &{Sky} &{Person} &{Car} &{Truck} &{Bus} &{Motor} &{Bike} &{mIoU} \\
     \shline
      Source only &Image &56.7 &18.7 &78.7 &6.0 &22.0 &15.6 &81.6 &18.3 &80.4 &59.9 &66.3 &4.5 &16.8 &20.4 &10.3 &37.1 \\
      AdvEnt~\cite{vu2019advent} &Image &78.5 &31.0 &81.5 &22.1 &29.2 &26.6 &81.8 &13.7 &80.5 &58.3 &64.0 &6.9 &38.4 &4.6 &1.3 &41.2 \\
      CBST~\cite{YangZou2018UnsupervisedDA} &Image &48.1 &20.2 &84.8 &12.0 &20.6 &19.2 &83.8 &18.4 &84.9 &59.2 &71.5 &3.2 &38.0 &23.8 &37.7 &41.7 \\
      IDA~\cite{pan2020unsupervised} &Image &78.7 &33.9 &82.3 &22.7 &28.5 &26.7 &82.5 &15.6 &79.7 &58.1 &64.2 &6.4 &41.2 &6.2 &3.1 &42.0 \\
      CRST~\cite{zou2019confidence} &Image &56.0 &23.1 &82.1 &11.6 &18.7 &17.2 &85.5 &17.5 &82.3 &60.8 &73.6 &3.6 &38.9 &30.5 &35.0 &42.4 \\
      FDA~\cite{yang2020fda} &Image &70.3 &27.7 &81.3 &17.6 &25.8 &20.0 &83.7 &31.3 &82.9 &57.1 &72.2 &22.4 &49.0 &17.2 &7.5 &44.4 \\
    \hline 
     DA-VSN~\cite{guan2021domain} &Video &86.8 &36.7 &83.5 &22.9 &30.2 &27.7 &83.6 &26.7 &80.3 &60.0 &79.1 &20.3 &47.2 &21.2 &11.4 &47.8 \\
     TPS~\cite{wu2022domain} &Video &82.4 &36.9 &79.5 &9.0 &26.3 &29.4 &78.5 &28.2 &81.8 &61.2 &80.2 &39.8 &40.3 &28.5 &31.7 &48.9 \\
     I2VDA~\cite{wu2022necessary} &Video &84.8 &36.1 &84.0 &28.0 &36.5 &36.0 &85.9 &32.5 &74.0 &63.2 &81.9 &33.0 &51.8 &39.9 &0.1 &51.2 \\
     STPL~\cite{lo2023spatio} &Video &83.1 &38.9 &81.9 &48.7 &32.7 &37.3 &84.4 &23.1 &64.4 &62.0 &82.1 &20.0 &76.4 &40.4 &12.8 &52.5 \\
     CMOM~\cite{cho2023domain} &Video &89.0 &53.8 &86.8 &31.0 &32.5 &47.3 &85.6 &25.1 &80.4 &65.1 &79.3 &21.6 &43.4 &25.7 &40.6 &53.8 \\
     \rowcolor{lightgray} 
     PiPa++ &Video &\textbf{94.0} &\textbf{61.8} &\textbf{89.8} &\textbf{41.4} &\textbf{40.5} &\textbf{55.3} &\textbf{90.6} &\textbf{37.3} &\textbf{85.3} &\textbf{69.0} &\textbf{82.3} &\textbf{51.0} &\textbf{59.6} &\textbf{28.7} &\textbf{48.6} &\textbf{62.3} \\
     \shline
    \end{tabular}}
    \label{table_viper}
    \end{table*}

    \begin{table*}[!ht]
        \caption{Quantitative comparison with previous video-based UDA methods on SYNTHIA-Seq~$\rightarrow$~Cityscapes-Seq. We present per-class IoU and mIoU. The best accuracy in every column is in bold.}
        \resizebox{\linewidth}{!}{
        \centering
        \begin{tabular}{c|c|ccccccccccc|c}
         \shline
         Method &Training &{Road} &{SW} &{Building} &{Pole} &{TL} &{TS} &{Veg.} &{Sky} &{Person} &{Rider} &{Car} &{mIoU} \\
         \shline
          Source only &Image &56.3 &26.6 &75.6 &25.5 &5.7 &15.6 &71.0 &58.5 &41.7 &17.1 &27.9 &38.3 \\
          AdvEnt~\cite{vu2019advent} &Image &85.7 &21.3 &70.9 &21.8 &4.8 &15.3 &59.5 &62.4 &46.8 &16.3 &64.6 &42.7 \\
          CBST~\cite{YangZou2018UnsupervisedDA} &Image &64.1 &30.5 &78.2 &28.9 &14.3 &21.3 &75.8 &62.6 &46.9 &20.2 &33.9 &43.3 \\
          IDA~\cite{pan2020unsupervised} &Image &87.0 &23.2 &71.3 &22.1 &4.1 &14.9 &58.8 &67.5 &45.2 &17.0 &73.4 &44.0 \\
          CRST~\cite{zou2019confidence} &Image &70.4 &31.4 &79.1 &27.6 &11.5 &20.7 &78.0 &67.2 &49.5 &17.1 &39.6 &44.7 \\
          FDA~\cite{yang2020fda} &Image &84.1 &32.8 &67.6 &28.1 &5.5 &20.3 &61.1 &64.8 &43.1 &19.0 &70.6 &45.2 \\
        \hline 
         DA-VSN~\cite{guan2021domain} &Video &89.4 &31.0 &77.4 &26.1 &9.1 &20.4 &75.4 &74.6 &42.9 &16.1 &82.4 &49.5 \\
         TPS~\cite{wu2022domain} &Video &91.2 &53.7 &74.9 &24.6 &17.9 &39.3 &68.1 &59.7 &57.2 &20.3 &84.5 &53.8 \\
         STPL~\cite{lo2023spatio} &Video &87.6 &42.5 &74.6 &27.7 &18.5 &35.9 &69.0 &55.5 &54.5 &17.5 &85.9 &51.8 \\
         I2VDA~\cite{wu2022necessary} &Video &89.9 &40.5 &77.6 &27.3 &18.7 &23.6 &76.1 &76.3 &48.5 &22.4 &82.1 &53.0 \\
         CMOM~\cite{cho2023domain} &Video &90.4 &39.2 &82.3 &30.2 &16.3 &29.6 &83.2 &84.9 &59.3 &19.7 &84.3 &56.3 \\
         \rowcolor{lightgray} 
         PiPa++ &Video &\textbf{94.4} &\textbf{67.2} &\textbf{89.6} &\textbf{47.5} &\textbf{58.8} &\textbf{61.0} &\textbf{88.9} &\textbf{89.5} &\textbf{68.3} &\textbf{42.5} &\textbf{87.7} &\textbf{72.3} \\
         \shline
        \end{tabular}}
        \label{table_video_syn}
        \end{table*}

\subsection{Ablation Studies and Further Analysis}

\noindent\textbf{Effect of Pixel-wise, Patch-wise, and Temporal Contrast.} 
We evaluate the effectiveness of the three primary components, \ie, Pixel-wise Contrast, Patch-wise Contrast and Temporal Contrast in the proposed PiPa++ and investigate how the combination of the contrasts contributes to the final performance.
In image-level DA, we explore the effect of Pixel-wise Contrast and Patch-wise Contrast on GTA $\rightarrow$ Cityscapes. 
For a fair comparison, we apply the same experimental settings and hyperparameters. 
We first reproduce the baseline DAFormer \cite{hoyer2022daformer}, which yields a competitive mIoU of 68.4. 
As shown in the Table~\ref{table:ablation}, we could observe: 
(1) Both Patch Contrast and Pixel Contrast individually could lead to +1.4 mIoU and +2.3 mIoU improvement respectively, verifying the effectiveness of exploring the inherent contextual knowledge. 
(2) The two kinds of contrasts are complementary to each other. The proposed method successfully mines the multi-level knowledge by combining the two kinds of contrast. When applying both losses, our PiPa further improves the network performance to 71.7 mIoU, surpassing the model that deploys only one kind of contrast by a clear margin. 
The second baseline model is HRDA \cite{hoyer2022hrda}. The observation is consistent with DAFormer. Using either pixel or patch loss could increase the performance, but jointly training them in a unified framework leads to the best results. Since HRDA introduces High Resolution (HR) and Low Resolution (LR) features, to effectively introduce Pixel-wise contrast and Patch-wise contrast in HRDA~\cite{hoyer2022hrda}, we conducted experiments on both HR and LR features as shown in Table~\ref{table:features}. It is shown that training with HR features results in higher performance. 
Further, in video-level DA, we explore the effect of PiPa as a combination and Temporal Contrast on VIPER $\rightarrow$ Cityscapes-Seq. It can be observed that implementing pixel-level and patch-level contrast on frames for video tasks boosts the performance from 53.2 to 58.8 mIoU. Furthermore, by using temporal contrast to aggregate temporal information, the performance reaches 62.3 mIoU, resulting in a total improvement of 9.1 mIoU.

\begin{table}[t]
    \centering
    \caption{Ablation study on the effect of Pixel-wise Contrast, Patch-wise Contrast, and Temporal Contrast based on two competitive baselines DAFormer\cite{hoyer2022daformer} and HRDA\cite{hoyer2022hrda}.}
    \label{table:ablation}
    \begin{tabular}{l|c|c|c|c|c}
    \shline
    Method & $\mathcal{L}_{\text{Pixel}}$ & $\mathcal{L}_{\text{Patch}}$ & $\mathcal{L}_{\text{Temp}}$ & mIoU & $\Delta$mIoU\\
    \hline 
    DAFormer\cite{hoyer2022daformer} & & & & $68.4$ & $-$ \\
    Patch Contrast &  & $\checkmark$ & & $69.8$ & +1.4 \\
    Pixel Contrast & $\checkmark$ & & & $70.7$ & +2.3 \\
    PiPa & $\checkmark$ & $\checkmark$ & & $71.7$ & +3.3 \\
    \shline
    HRDA\cite{hoyer2022hrda} & & & & $73.8$ & $-$ \\
    Patch Contrast &  & $\checkmark$ & & $74.7$ & +0.9 \\
    Pixel Contrast & $\checkmark$ & & & $74.9$ & +1.1 \\
    PiPa & $\checkmark$ & $\checkmark$ & & $75.6$ & +1.8 \\
    \hline
    Video Baseline &  &  &  & $53.2$ &  \\
    PiPa & $\checkmark$ & $\checkmark$ &  & $58.8$ & +5.6 \\
    PiPa++ & $\checkmark$ & $\checkmark$ & $\checkmark$ & $62.3$ & +9.1 \\
    \shline
    \end{tabular}
\end{table}

\begin{table}[t]
    \centering
    \caption{Sensitivity analysis of the pseudo label threshold.}
    \label{table:threshold} 
    \begin{tabular}{c|c|c|c|c|c|c|c}
    \shline Threshold & 0.6 & 0.7 & 0.8 & 0.9 & 0.95 & 0.968 & 0.99 \\
    \hline 
    $\mathrm{mIoU}$  & $66.3$ & $68.9$ & $69.4$ & $70.8$ & $71.2$ & $71.7$ & $71.4$ \\
    \shline
    \end{tabular}
\end{table}

\noindent\textbf{Effect of Task-smart Sampling.}
As shown in Table~\ref{table:sampling}, PiPa++ firstly constructed an inside-image representation space for image-level tasks, achieving a performance of 75.3 mIoU. Subsequently, PiPa++ established a memory bank to accommodate more training samples. When expanded to the whole training batch, it reached 75.6 mIoU, and when further extended to the entire training dataset, it achieved 76.6 mIoU. This validates the effectiveness of having more contrastive samples. In addition, PiPa++ tested two sampling strategies, Long Range Sampling and Short Range Sampling, for video-level tasks on the VIPER dataset. It can be observed that the performance decreases when PiPa++ selects samples from frames that are farther apart.

\begin{table}[t]
    \centering
    \caption{Effect of task-smart sampling.}
    \label{table:sampling}
    \begin{tabular}{c|c|c|c}
    \shline Sampling Type & Sampling Strategies & $\mathrm{mIoU}$ & $\Delta \mathrm{mIoU}$ \\
    \hline
    \multirow{3}{*}{Image-based} & Image Samp. & $75.3$ & $-$ \\
                                 & Whole Batch Samp. & $75.6$ & $+0.3$ \\
                                 & Entire Dataset Samp. (PiPa++) & $76.6$ & $+1.3$ \\
    \hline
    \multirow{2}{*}{Video-based} & Long Range Samp. & $60.9$ & $-$ \\
                                 & Short Range Samp. & $62.3$ & $+1.4$ \\
    \shline
    \end{tabular}
\end{table}

\noindent\textbf{Effect of the patch crop size.}
For the patch contrast, the size of the patch also affects the number of negative pixels and training difficulty. As shown in Table~\ref{table:crop}, we gradually increase the patch size. We observe that larger patch generally obtain better performance since it contains more diverse contexts. 
There are two main advantages when increasing the patch size: (1) In larger patches, we could include more ``hard negative'' pixels for contrastive learning; 
(2) In larger patches, we have a larger receptive field, which could include contextual cues for bigger objects, such as trains.
It is also worth noting that if the patch size is too large (like 960), the overlapping area can be larger than the non-overlapping area, which also may compromise the training.  
\begin{table}[t]
    \centering
    \caption{Effect of different crop types in HRDA~\cite{hoyer2022hrda}.}
    \label{table:features}
    \vspace{+2pt}
    \begin{tabular}{c|c}
    \shline Method & mIoU \\
    \hline 
    LR Crops  & $75.1$  \\
    HR Crops  & $75.6$  \\
    \shline
    \end{tabular}
    \vspace{+1pt}
    \end{table}
    
    \begin{table}[t]
    \centering
    \caption{Effect of the patch crop size.}
    \label{table:crop}
    \begin{tabular}{c|c}
    \shline Crop Size & $\mathrm{mIoU}$ \\
    \hline 
    $480 \times 480$  & $70.4$ \\
    $600 \times 600$  & $71.0$ \\
    $720 \times 720$  & $71.7$ \\
    $900 \times 900$  & $70.9$ \\
    \shline
    \end{tabular}
    \end{table}

\noindent\textbf{Sensitivity of the pseudo label threshold.}
Since the target annotation is not available in unsupervised domain adaptation, a hard threshold beta is used to eliminate low-confidence pixel predictions from the predicted label. We conducted additional experiments on the threshold and found that within the range of 0.9-0.99, the DAFormer + PiPa results were not sensitive to the beta in Table~\ref{table:threshold}. We set the threshold to 0.968 to obtain optimal results following previous self-training works \cite{WilhelmTranheden2020DACSDA, hoyer2022daformer}.

\begin{table}[t]
    \centering
    \caption{Ablation study on video-level domain adaptation using PiPa and PiPa++ on the Synthia Seq dataset.}
    \label{table:ablation_study}
    \begin{tabular}{c|c}
    \shline
    Method & mIoU \\
    \hline
    Baseline (no adaptation) & 60.5 \\
    \hline
    PiPa & 65.7 \\
    \hline
    PiPa++ & 68.3 \\
    \shline
    \end{tabular}
    \end{table}

\noindent\textbf{Multi source domain setting.}
By incorporating multi-source domain data, the model can be trained to be more robust to the unlabelled target environment~\cite{zhao2019multi, he2021multi}. We first adopt previous work MADAN~\cite{zhao2019multi} as our baseline, which reaches 41.4 mIoU on GTA5 + SYNTHIA $\rightarrow$ Cityscapes. MADAN + PiPa increases the performance to 44.1 mIoU. Then we adopt a self-training baseline DACS \cite{WilhelmTranheden2020DACSDA}, which achieves a mIoU of 52.1 (Only GTA) as shown in Table~\ref{table:multisource}. By incorporating additional source-domain data, the model's performance improves to 54.2 mIoU. Our proposed method further improves the model's performance, increasing the mIoU from 54.2 to 56.1 mIoU, demonstrating consistent improvement over various baselines.

\begin{table}[t]
    \centering
    \caption{Results on GTA5 + SYNTHIA $\rightarrow$ Cityscapes.}
    \label{table:multisource}
    \begin{tabular}{c|c|c}
    \shline 
    Base & Multi Src. &  Multi Src + PiPa  \\
    \hline
    52.1 & 54.2 & 56.1 \\
    \shline
    \end{tabular}
\end{table}

\noindent\textbf{Ablation study on Normal-to-Adverse setting.}
\noindent
ACDC is a large dataset with 4,006 images containing four common adverse conditions: fog, nighttime, rain and snow. In Cityscapes $\rightarrow$ ACDC, the knowledge is transferred from the source domain under normal visual conditions, \ie, at daytime and in clear weather to adverse visual conditions. The quantitative comparisons are shown in Table~\ref{table:acdc}. We can observe that our PiPa yields a significant improvement over the previous methods. Particularly, PiPa achieves 58.6 mIoU, which outperforms DAFormer by +3.2 mIoU, which demonstrates the competitive generalization ability of PiPa in adverse visual conditions. When plugging on recent works MIC~\cite{hoyer2023mic} and Refign~\cite{bruggemann2023refign}, PiPa shows consistent improvement.

\begin{table}[]
    \centering
    \caption{Quantitative comparison with previous UDA methods on Cityscapes $\rightarrow$ ACDC. The performance is provided as mIoU in $\%$ and the best result is in \textbf{bold}.}
    \label{table:acdc} \small
    \begin{tabular}{l|c|c}
    \shline 
    Method & Architecture & mIoU \\
    \hline
    ADVENT \cite{vu2019advent} & DeepLabv2 & 32.7 \\
    AdaptSegNet \cite{tsai2018learning} & DeepLabv2 & 32.7 \\
    BDL \cite{li2019bidirectional} & DeepLabv2 & 37.7 \\
    CLAN \cite{luo2019taking} & DeepLabv2 & 39.0 \\
    FDA \cite{yang2020fda} & DeepLabv2 & 45.7 \\
    MGCDA \cite{sakaridis2020map} & DeepLabv2 & 48.7 \\
    DANNet \cite{wu2021dannet} & DeepLabv2 & 50.0 \\
    \hline
    DAFormer \cite{hoyer2022daformer} & Transformer & 55.4 \\
    \rowcolor{lightgray} 
    DAFormer \cite{hoyer2022daformer} + PiPa & Transformer & \textbf{58.6 (+3.2)}\\
    MIC \cite{hoyer2023mic} & Transformer & 59.2 \\
    \rowcolor{lightgray} 
    MIC \cite{hoyer2023mic} + PiPa & Transformer & \textbf{61.1 (+1.9)}\\
    Refign \cite{bruggemann2023refign} & Transformer & 65.5 \\
    \rowcolor{lightgray} 
    Refign \cite{bruggemann2023refign} + PiPa & Transformer & \textbf{66.4 (+0.9)}\\
    \shline
    \end{tabular}
    \vspace{-0.2in}
\end{table}

\begin{table}[]
    \centering
    \caption{Quantitative Results on Cityscapes $\rightarrow$ Oxford-Robot \cite{maddern20171}. The performance is provided as mIoU in $\%$ and the best result is in \textbf{bold}.}
    \label{table:oxford} \tiny
    \begin{tabular}{l | c c c c c c c c c | c}
    \shline
    Method & \rotatebox{90}{road} & \rotatebox{90}{sidewalk} & \rotatebox{90}{building} & \rotatebox{90}{light} & \rotatebox{90}{sign} & \rotatebox{90}{sky} & \rotatebox{90}{person} & \rotatebox{90}{automobile} & \rotatebox{90}{two-wheel} & mIoU \\
    \hline
    MRNet \cite{zheng2019unsupervised} & 95.9 & 73.5 & 86.2 & 69.3 & 31.9 & 87.3 & 57.9 & 88.8 & 61.5 & 72.5 \\
    \rowcolor{lightgray} 
    MRNet + PiPa & \textbf{96.9} & 75.1 & 88.0 & 69.9 & 36.5 & 88.8 & 61.5 & 89.1 & \textbf{63.1} & 74.3 \\
    \hline 
    Uncertainty \cite{zheng2021rectifying} & 95.9 & 73.7 & 87.4 & 72.8 & \textbf{43.1} & 88.6 & 61.7 & 89.6 & 57.0 & 74.4 \\
    \rowcolor{lightgray} 
    Uncertainty + PiPa & 96.0 & \textbf{76.2} & \textbf{93.3} & \textbf{73.3} & 42.5 & \textbf{90.9} & \textbf{65.4} & \textbf{91.1} & 59.5 & \textbf{76.5} \\
    \shline
    \end{tabular}
\end{table}

\noindent
Oxford RobotCar dataset \cite{maddern20171} contains 894 training images with 9 classes and is collected during rainy and cloudy weather conditions, presenting a challenge due to the noisy variants introduced by such illumination conditions. We observe that the proposed method also has achieved the competitive results on Cityscapes $\rightarrow$ Oxford-Robot based on MRNet \cite{zheng2019unsupervised} and Uncertainty \cite{zheng2021rectifying}, reaching 1.8 and 2.1 mIoU increase respectively.

\section{Conclusion and Further Discussions}
\label{sec: conclution}
In this paper, we presented PiPa++, a unified framework for unsupervised domain adaptive semantic segmentation that leverages self-supervised learning techniques to bridge the gap between image- and video-level domain adaptation. Our approach integrates pixel-wise, patch-wise, and temporal contrastive learning to capture both spatial and temporal contextual information, enabling robust feature representation and improved segmentation performance across different domains.

The extensive experiments conducted on both image-based and video-based UDA benchmarks demonstrate the effectiveness and versatility of PiPa++. For image-based tasks, PiPa++ achieves significant performance gains by constructing an inside-image representation space and extending it to the entire training dataset through a semantic-aware memory bank. For video-based tasks, the proposed task-smart sampling strategy ensures the selection of informative samples, maintaining temporal consistency and enhancing segmentation accuracy.

Future work will focus on exploring more advanced contrastive learning techniques and further enhancing the scalability of PiPa++ to handle larger and more diverse datasets. Additionally, we plan to investigate the integration of PiPa++ with other vision tasks, such as object detection and instance segmentation, to develop a more comprehensive and unified framework for visual understanding.

\bibliographystyle{IEEEtran}
\bibliography{IEEEabrv,ref}

\begin{IEEEbiography}[{\includegraphics[width=1in,height=1.25in,clip,keepaspectratio]{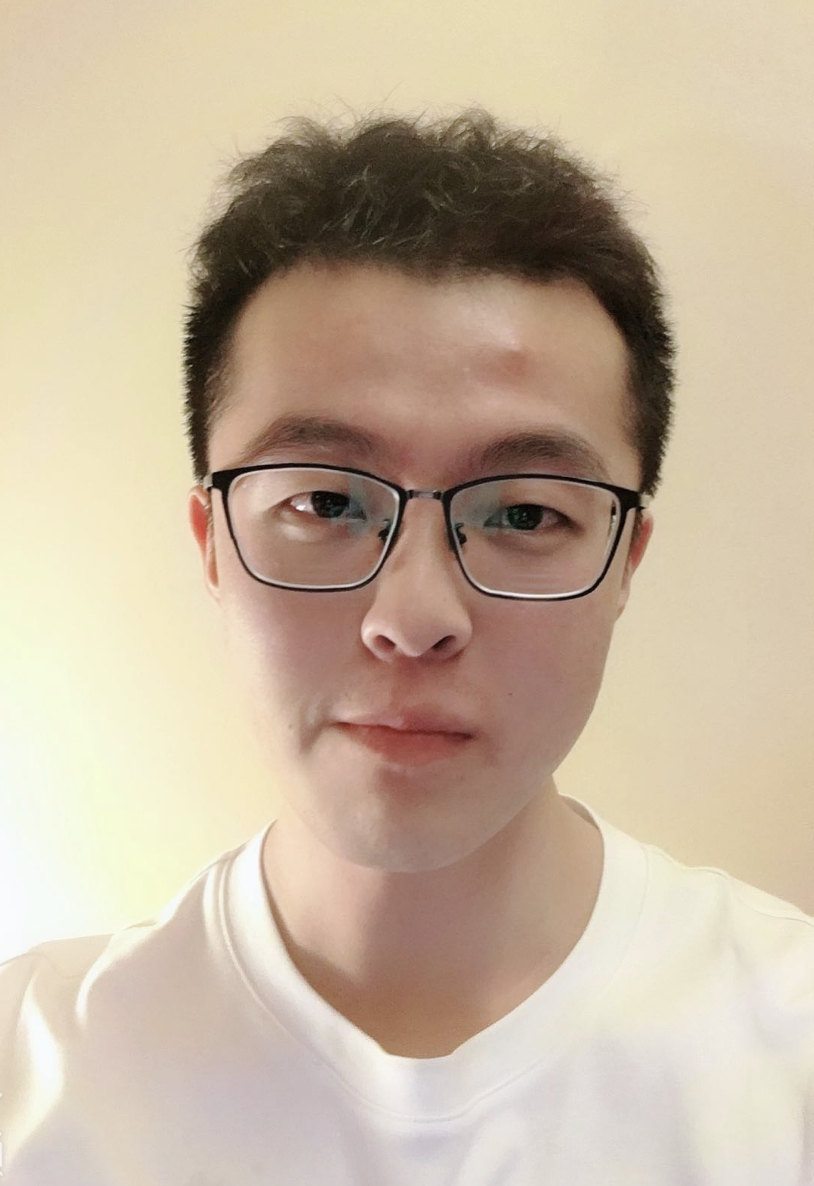}}]{Mu Chen}
    received the B.E. degree from the school of Engineering at Monash University, Australia in 2021. He is currently pursuing a
    PhD degree at ReLER Lab, Australian Artificial Intelligence Institute, University of Technology Sydney, Australia. His research interests include
    Unsupervised Domain Adaptation, Image Segmentation and Video Segmentation.
\end{IEEEbiography}

\begin{IEEEbiography}[{\includegraphics[width=1in,height=1.25in,clip,keepaspectratio]{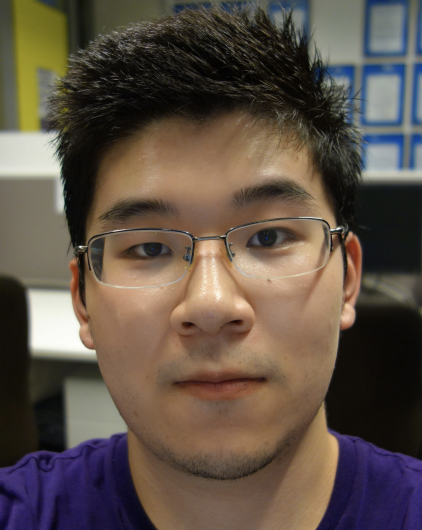}}]{Zhedong Zheng}
is an assistant professor with the University of Macau. He was a research fellow at School of Computing, National University of Singapore. He received the Ph.D. degree from the University of Technology Sydney, Australia, in 2021 and the B.S. degree from Fudan University, China, in 2016.
He received the IEEE Circuits and Systems Society Outstanding Young Author Award of 2021. 
He has 
organized a special session on reliable retrieval at ICME'22, two workshops at ACM MM'23 and one workshop at ACM ICMR'24.
Besides, he was invited as a keynote speaker at CVPR'20, CVPR'21, a tutorial speaker at ACM MM'22. He also serves as an area chair at ACM MM'24.
\end{IEEEbiography}

\begin{IEEEbiography}
[{\includegraphics[width=1in,height=1.25in,clip,keepaspectratio]{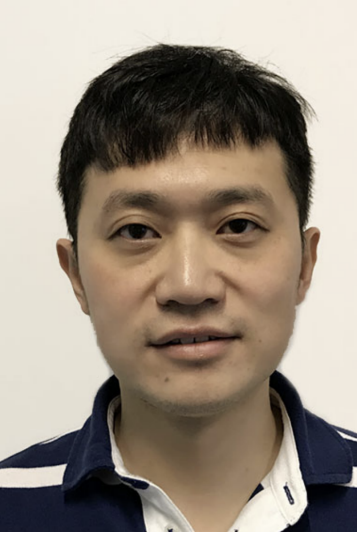}}]{Yi Yang}
 (Senior Member, IEEE) received the PhD
degree from Zhejiang University, in 2010. He is
a distinguished professor at Zhejiang University,
China. His current research interests include machine learning and multimedia content analysis, such
as multimedia retrieval and video content understanding. He received the Australia Research Council Early Career Researcher Award, the Australia
Computing Society, the Google Faculty Research
Award, and the AWS Machine Learning Research
Award Gold Disruptor Award.
\end{IEEEbiography}

\end{document}